# Deep Learning for identifying radiogenomic associations in breast cancer

Zhe Zhu[a,*], Ehab Albadawy[a], Ashirbani Saha[a], Jun Zhang[a], Michael R. Harowicz[a], Maciej A. Mazurowski[a,b]

[a]Department of Radiology, Duke University, Durham, NC 27705, USA
[b]Department of Electrical and Computer Engineering, Duke University, Durham, NC 27708, USA
[*]Corresponding author.

E-mail address: zhe.zhu@duke.edu(Z.Zhu), ehalbadawy93@gmail.com(E. Albadawy),
as698@duke.edu(A.Saha), xdzhangjun@gmail.com(J.Zhang),
michael.harowicz@gmail.com(M.Harrowicz), maciej.mazurowski@duke.edu(M.Mazurowski)

## Abstract

**Purpose**: To determine whether deep learning models can distinguish between breast cancer molecular subtypes based on dynamic contrast-enhanced magnetic resonance imaging (DCE-MRI).

**Materials and methods:** In this institutional review board–approved single-center study, we analyzed DCE-MR images of 270 patients at our institution. Lesions of interest were identified by radiologists. The task was to automatically determine whether the tumor is of the Luminal A subtype or of another subtype based on the MR image patches representing the tumor. Three different deep learning approaches were used to classify the tumor according to their molecular subtypes: learning from scratch where only tumor patches were used for training, transfer learning where networks pre-trained on natural images were fine-tuned using tumor patches, and off-the-shelf deep features where the features extracted by neural networks trained on natural images were used for classification with a support vector machine. Network architectures utilized in our experiments were GoogleNet, VGG, and CIFAR. We used 10-fold crossvalidation method for validation and area under the receiver operating characteristic (AUC) as the measure of performance.

**Results:** The best AUC performance for distinguishing molecular subtypes was 0.65 (95% CI:[0.57,0.71]) and was achieved by the off-the-shelf deep features approach. The highest AUC performance for training from scratch was 0.58 (95% CI:[0.51,0.64]) and the best AUC performance for transfer learning was 0.60 (95% CI:[0.52,0.65]) respectively. For the off-the-shelf approach, the features extracted from the fully connected layer performed the best.

**Conclusion:** Deep learning may play a role in discovering radiogenomic associations in breast cancer.

## Introduction

Molecular classification into intrinsic subtypes has led to significant advances in the field of breast cancer. Each distinct molecular subtype is associated with a tendency of disease progression, which is why treatment recommendations from physicians hinge on the genomic analysis of each patient's tumor. Recently, the field of radiogenomics or imaging genomics has emerged, which aims at finding correlations between the imaging characteristics of cancer and its genomic composition. A specific area that has garnered significant attention is the prediction of genomics in breast cancer using MRI[1–16]. Previous work on this topic utilized either imaging features manually extracted by radiologists, which is a very time consuming and subjective process, or features automatically extracted by computer algorithms. Such features include tumor texture (e.g. Haralick features), tumor shape, or enhancement dynamics[6]. While this has shown promising results, hand-crafted features are limited because they require the researcher to anticipate which characteristics might be of use for a given tumor.

Deep learning approaches have shown superiority to handcrafted approaches in automatic feature extraction[17], image classification[18] and object detection[19]. In this study we propose using deep learning to conduct radiogenomic analysis of breast cancer. Specifically, we studied three different deep learning approaches: training from scratch, transfer learning and off-the-shelf deep features. Training from scratch is the most straightforward way of training deep neural networks when there are enough training data. In many medical imaging tasks, however, the number of training samples is insufficient for this method. In these situations using the training from scratch approach can cause overfitting. One way to alleviate the issue of limited data is to use transfer learning. The transfer learning approach initializes the network using a model pre-trained using different data (e.g., natural images) and then additionally trains the network using the specific data for the task at hand. Another option when faced with limited data is to use the deep features approach, which utilizes the pre-trained network as the feature extractor. Afterwards a traditional classifier such as a support vector machine is trained on the extracted features. Each of these approaches have been proven to work well on a specific subset of medical imaging tasks[20–22].

## Materials and Methods
### Patient Population
Design and execution of this study was approved by institutional review board. We collected consecutive preoperative dynamic contrast enhancement MRIs of 400 patients at Duke University Medical Center acquired from September 2007 to June 2009. Then, we excluded 114 patients for the following reasons: 19 had a previous history of breast cancer, 19 had a history of benign elective breast surgery, 29 were undergoing breast cancer treatment at the time of MRI, 42 had missing pathology data, 3 had missing sequences, 1 had a discordant number of slices in pre-contrast and post-contrast sequences, 1 had no biopsy performed. The remaining 286 cases were split into 6 subsets without overlap and were annotated by six breast imagers. 11 cases were skipped by these readers (reasons include: tumor was not very clear in the MRIs and the reader was not confident for the cases). For the remaining 275 cases, 3 had errors in image processing, 1 had implants and 1 was missing the post-contrast sequences. The remaining 270 cases were used for our study. 90 cases belonged to luminal A subtype while the rest belonged to the other 3 subtypes.

### Imaging and Pathology Data

All MRIs in this study were acquired using a 1.5 Tesla (Signa HDx, GE Healthcare, Little Chalfont, United Kingdom[44]; Signa HDxt, GE Healthcare[5]; MAGNETOM Avanto, Siemens, Munich, Germany[37]) or 3.0 Tesla scanner (Signa HDx, GE Healthcare[167]; MAGNETOM Trio, Siemens, Munich, Germany[25]) scanner using a breast coil (Invivo, Orlando, FL). Each case had the following sequences: nonfat-saturated T1-weighted, fat-saturated T2-weighted sequence, and pre-contrast followed by three dynamic post-contrast T1-weighted gradient echo series with fat suppression after intravenous administration of gadopentetate dimeglumine (Magnevist, Bayer Health Care, Berlin, Germany) or gadobenate dimeglumine (MultiHance, Bracco, Milan, Italy). Contrast load was determined using a weight based dosing protocol (0.2 mL/kg). The estrogen receptor (ER), progesterone receptor (PR), and HER2-neu status were obtained from the initial breast biopsy pathology report.

### Image Annotation
Six fellowship-trained breast-imaging radiologists with 6–20 years of experience finished the annotation of the dataset. For each case up to 5 bounding boxes were annotated by the reader. The bounding boxes represented a lesion(s). Finally, the annotated bounding boxes were converted to binary masks where 1 indicates inside the bounding box and 0 indicates outside of the bounding box.

### MR Imaging Pre-processing and Tumor Patch Extraction
The MRIs in our dataset contain differences in pixel spacing as they were captured by different devices. Thus, the first step was to register all the MRIs to the same spatial resolution. This was achieved by accounting for the frequency of different spatial resolutions, and selecting the resolution that occurred most often as the target resolution. Next, MRIs with other spatial resolutions were scaled to the target resolution using bilinear interpolation. Three channel images were constructed by concatenating $T' - T^p, T'' - T^p, T''' - T^p$, where $T'$, $T''$ and $T'''$ are three post-contrast sequences and $T^p$ is the pre-contrast sequence. We assumed that the center of the bounding box is the center of the lesion, and we sampled square patches around the lesion center. The patch size was altered between 80 pixels and 120 pixels in our experiments. For each lesion, four additional patches were generated by random translation and rotation, and 44660 patches were generated in total for a given patch size.

### Methods Overview
Three different deep learning approaches were used in our experiment: training from scratch, transfer learning and off-the-shelf deep features approach. Pipelines of these approaches are illustrated Figure 1. We used CAFFE[23] deep learning framework on a desktop with a NVIDIA GTX 1080 GPU. In our experiment we trained the models for 40 epochs.

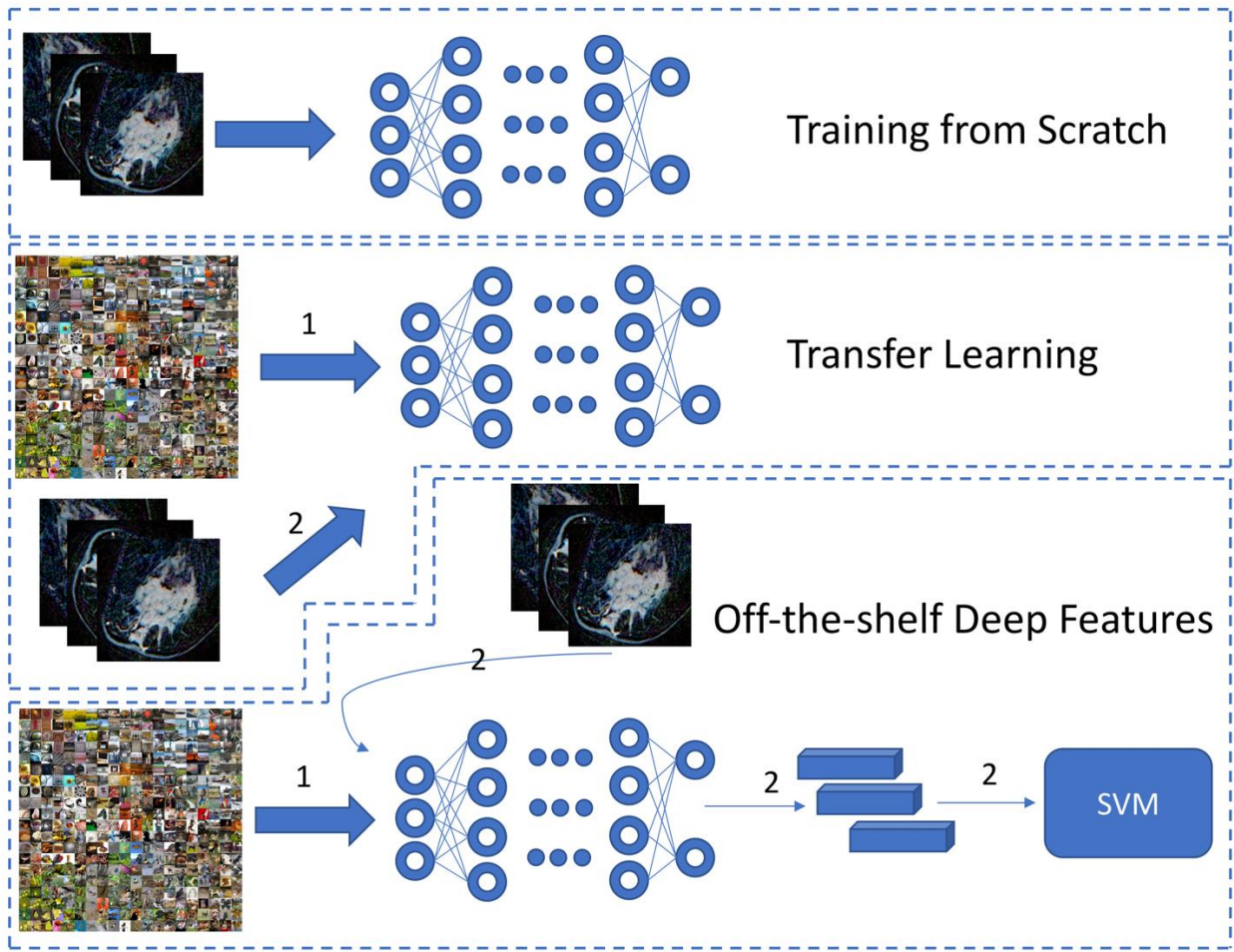

Figure 1. Pipeline of the three approaches. Training from scratch: directly training a neural network using our dataset. Transfer learning: pre-training a model on natural image dataset, then fine-tuning it using our dataset. Off-the-shelf deep features: using pre-trained neural network as feature extractor, and then training a traditional classifier(SVM) using the extracted features. Thick arrows indicate training process while thin arrows indicate feeding images to neural networks.

**Training from Scratch**
Training from scratch is the most straightforward way to train convolutional neural networks. In this approach, the network is initialized with random weights and the training is performed with the available data. There are two factors that greatly affect the performance: training dataset and network architecture. Current neural networks contain tens or even hundreds of layers, resulting in millions of free parameters to train. Building a large dataset requires large amount of time and labor, and in some situations such as in medical imaging the amount of available data is often insufficient. Since improving the quality of training dataset is challenging, more attention has been paid to the improvement of network structure. Recently many neural network structures[24,25] have been proposed which perform well on a variety of different tasks. We choose three representative network architectures in our study: GoogleNet[24], VGGNet[25] and CIFAR[26]. For GoogleNet we used both original GoogleNet and reduced GoogleNet while for VGGNet we just used the reduced VGGNet, resulting in four specific architectures in total. In our problem the lesions on MRI have different sizes which requires the network to have multi-scale capability. We chose GoogleNet for this reason. The main building block of GoogleNet is the Inception module that consists of three convolution layers with different kernel sizes. These kernels can be used to capture the features of patterns of different sizes. VGGNet is much larger than GoogleNet, which means it has more weights and needs more data to train. Thus we only used reduced VGGNet due to the data insufficiency. For the original GoogleNet a much smaller learning rate (0.0001) than default (0.01) was chosen, as image patches in our dataset had far less variation compared with those natural images. For convenience, we refer the reduced GoogleNet as GoogleNet_R and reduced VGGNet as VGG_R. To reduce the GoogleNet, in the first convolution layer we decreased the number of filters from 64 to 32, the pad size from 3 to 2, the kernel size from 7 to 5, the stride from 2 to 1 and the stride of the subsequent pooling layer from 2 to 1. The input size of the network was fixed to 64×64×3. To reduce VGGNet, we only retained the first 2 convolution layers and all the fully connected layers, and input size of the neural network was fixed to 80×80×3. The output dimension of the last layer of these networks were modified from 1000 to 2, to adapt to the binary classification setting. For CIFAR network, we set the input size to 80×80×3 and the output dimension of the last layer to 2.

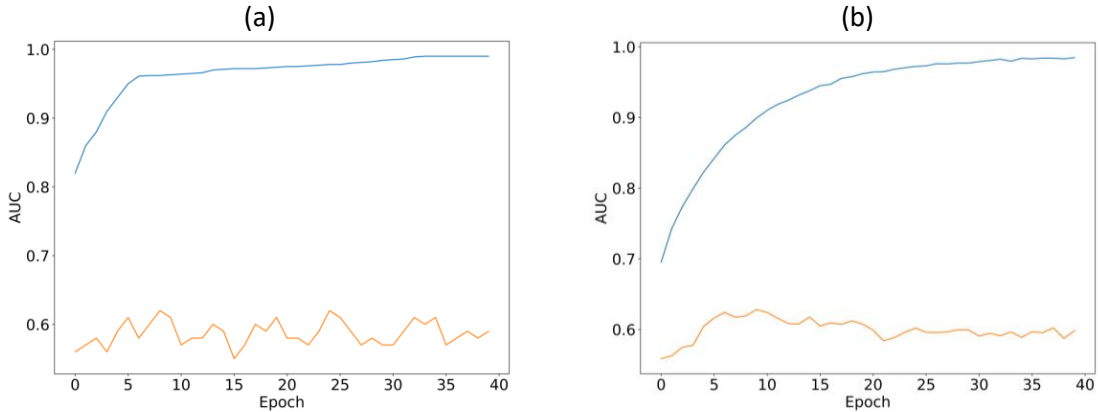

Figure 2. (a) AUCs of training from scratch for reduced VGGNet. Orange line: test AUC; blue line: training AUC. (b) AUCs of transfer learning using pre-trained GoogleNet.

## Transfer Learning

Transfer learning is a two-step approach. The first step is to pre-train the network using a large dataset for a different task than the one at hand. The second step is fine-tuning the pre-trained network on the dataset representing the problem of interest (a.k.a., the target dataset). Usually the target dataset is much smaller than the dataset used for pre-training. In our scenario we chose natural image dataset for pre-training and fine-tuned the pre-trained network on our own dataset. The networks pre-trained on large natural image datasets are already publicly available. We chose GoogleNet and VGGNet pre-trained on ImageNet for transfer learning. Following modifications were made to adapt the networks to our specific problem: the output dimension of the last fully connected layer was changed from 1000 to 2, and the weights in this layer were re-initialized randomly. The first modification is intuitive as we are dealing with a binary classification problem. The second modification is based on the observation[27] that last fully connected layer is the most problem-specific layer, so it needs to be completely retrained. The learning rate of the last fully connected layer was set to 0.001, which is 10 times larger than the learning rate of all the other layers (0.0001). When pre-trained on ImageNet, the input resolution of GoogleNet and VGGNet were fixed to 224×224×3. To be consistent with this resolution, all the patches were resized to 256×256×3. Then a 224×224×3 sub-patch was randomly cropped from the 256×256×3 image in each epoch.

## Off-the-shelf deep features

While training from scratch and transfer learning are both end-to-end approaches, off-the-shelf deep features approach explicitly separate feature extraction and classification. Since this approach only uses pre-trained networks as feature extractors and trains traditional classifiers using the extracted features, it can avoid the overfitting while taking advantage of the richness of deep features. Another benefit of using deep features approach is that the extracted features can be used for many different tasks. For example pre-trained network was originally used for image classification, but when used as feature extractor the extracted features were used for scene recognition[17]. We chose GoogleNet and VGGNet pre-trained on ImageNet as feature extractors and trained SVMs using the extracted features. While deep neural networks have multiple layers, it remains a question which layer's output should be used as features. The shallow layers may output low level but less specific features. Feature maps of deeper layers are higher level features but are also more problem-specific. We followed a previously described approach[17] and used the feature map of the last fully connected layer as default. Other layers' feature maps were also evaluated for comparison. We max pooled the feature map on the $x - y$ coordinates of the image plane, in order to limit the length of the feature vector.

For our study there is a limited number of training samples, and SVM would not perform well when its feature length becomes larger than its training samples. SVMs with different kernel functions were then trained and compared.

## Model Evaluation

Our patch-based approach treated each patch as a training sample in training phase. In test phase each patient was regarded as a test case. Five consecutive slices that contained the largest size of lesions were chosen and for each slice five patches (four corners and middle region of the lesion) were sampled. The averaged score of the 25 patches was used as the final score.

We separated our dataset into 10 folds where each fold had approximately the same ratio between the number of positive samples and the number of negative samples. The performance of each fold was evaluated separately and the final AUC was obtained by averaging the AUC of the 10 folds. The bootstrapping strategy was used to calculate the confidence interval.

## Results

The results for the models trained from scratch are illustrated in Table 1. Regarding the performance on training set (third column in Table 1), the AUC of GoogleNet, reduced GoogleNet and reduced VGGNet were all above 0.90, indicating that those trained networks fitted the training data distribution well. Note that even the reduced version of GoogleNet and VGGNet were still large networks. For CIFAR which is a rather small network, the AUC on training set was much lower than those large networks, but it had the highest AUC on test set. This indicates that larger networks are more likely to overfit when trained on a small dataset. The 95% confidence intervals of the test AUCs of the 4 networks were [0.51,0.60], [0.51,0.62], [0.46,0.60], and [0.52,0.63] respectively. AUCs of reduced VGGNet of the training and test set during training phase are plotted in Figure 2a. While the performance of reduced VGGNet on the training set reached high level quickly, on test set it stayed around the 0.58 for the remainder of the training.

Table 1. Results of training from scratch. We show the results of four different networks: original GoogleNet, reduced GoogleNet(GoogleNet_R), reduced VGGNet(VGG_R) and CIFAR. We list network input, training AUC and test AUC.

|  | Network Input | Training AUC | Test AUC |
|---|---|---|---|
| GoogleNet | 224×224×3 | 0.93 | 0.56 |
| GoogleNet_R | 64×64×3 | 0.94 | 0.57 |
| VGGNet_R | 80×80×3 | 0.98 | 0.54 |
| CIFAR | 80×80×3 | 0.75 | 0.58 |

The results of transfer learning are shown in Table 2. Regarding the test AUC, transfer learning always performed better than training from scratch. We also plot the AUCs of the training and test set of GoogleNet during training phase in Figure 2(b). Compared with the curves in Figure 2(a), the training AUC curve in Figure 2(b) reaches its asymptote more slowly than 2(a) while the test AUC curve in Figure 2(b) stays higher. This indicates that transfer learning can better avoid overfitting than training from scratch. It also indicates that pre-trained neural networks on large dataset can adapted to small dataset even when there exist large style differences between the two datasets. The 95% confidence intervals of the test AUCs of the 2 networks were [0.55,0.66] and [0.52,0.65].

Table 2. Results of transfer learning. We choose ImageNet pre-trained GoogleNet and VGGNet and perform transfer learning on our own dataset.

|  | Network Input | Training AUC | Test AUC |
|---|---|---|---|
| GoogleNet | 224×224×3 | 0.93 | 0.60 |
| VGGNet | 224×224×3 | 0.89 | 0.59 |

For off-the-shelf deep features approach both patch size in the feature extraction and kernel function in the SVM affect the final performance. We tested several combinations and the results are shown in Table 3. We found that using GoogleNet with 120 patch size and polynomial kernel SVM achieved best performance. Results using other kernel functions(rbf kernel and linear kernel) in SVM, other patch size(80) and other feature extraction network(VGGNet) are also given. From the third, fourth and fifth row we can infer that polynomial kernel best suits our problem. Two results of VGGNet are given in sixth and seventh row. Compared with the corresponding results using GoogleNet but the same patch size and kernel function (row 2 and row 4), GoogleNet performed better than VGGNet on our dataset. The 95% confidence intervals of the test AUCs of the 6 results were [0.51,0.64], [0.51,0.65], [0.57,0.71], [0.50,0.60], [0.50,0.59] and [0.48,0.62] respectively.

Table 3. AUCs with different patch sizes and kernels of SVM.

| Network | Patch Size | Kernel Function | Training AUC | Test AUC |
|---|---|---|---|---|
| GoogleNet | 80 | poly | 0.82 | 0.58 |
| GoogleNet | 120 | rbf | 0.85 | 0.59 |
| GoogleNet | 120 | poly | 0.84 | 0.65 |
| GoogleNet | 120 | linear | 0.74 | 0.56 |
| VGGNet | 80 | poly | 0.86 | 0.55 |
| VGGNet | 120 | poly | 0.83 | 0.56 |

We also tested the performance of features extracted from different layers, including intermediate convolution layers and fully connected layers. The performance of each layer's feature map is shown in Figure 3 and Table 4. Regarding the test AUC, feature map of the last fully connected layer has the best performance.

Table 4. Performance of features extracted from different layers.

|  | Conv1 | Conv1 | Incep1 | Incep2 | Incep3 | Incep4 | Incep5 | Incep6 | Incep7 | Incep8 | Incep9 | FC1 |
|---|---|---|---|---|---|---|---|---|---|---|---|---|
| Feature Length | 64 | 192 | 256 | 480 | 512 | 512 | 512 | 528 | 832 | 832 | 1024 | 1000 |
| Training AUC | 0.80 | 0.90 | 0.94 | 0.93 | 0.90 | 0.86 | 0.88 | 0.88 | 0.91 | 0.88 | 0.89 | 0.84 |
| Test AUC | 0.60 | 0.63 | 0.61 | 0.59 | 0.58 | 0.52 | 0.54 | 0.56 | 0.58 | 0.58 | 0.60 | 0.65 |

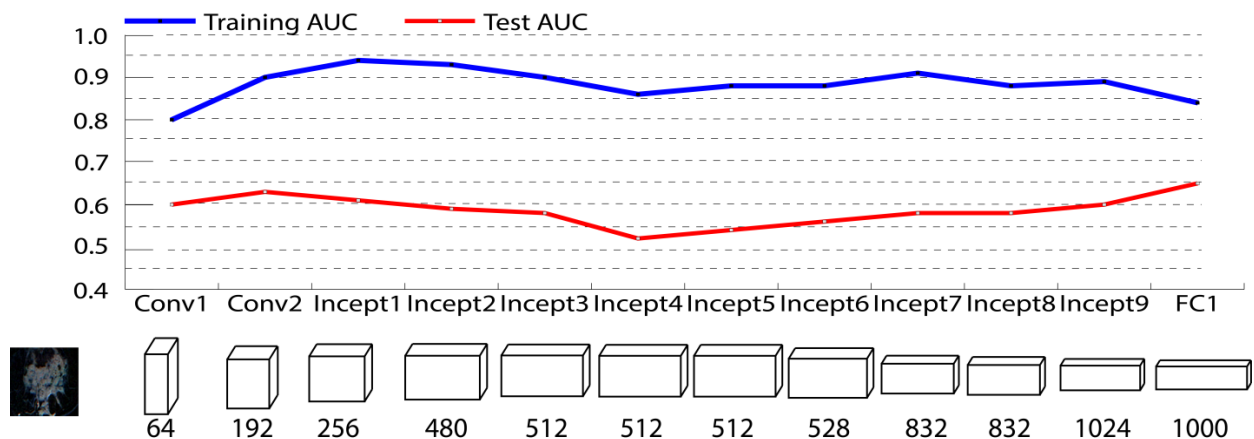

Figure 3. AUCs of different layer's feature map of pre-trained GoogleNet.

To investigate the performance of each individual feature, we used each single feature of the final fully connected layer's feature map of GoogleNet as the score for prediction. AUCs of those 1000 features were calculated and ranked, and the results are shown in Figure 4. The best AUC using a single feature is 0.64 which is slightly worse than using all the features with a polynomial kernel SVM. Please note that these are not crossvalidation results and it is an evaluation of 1000 features and therefore there is some bias to this performance estimate.

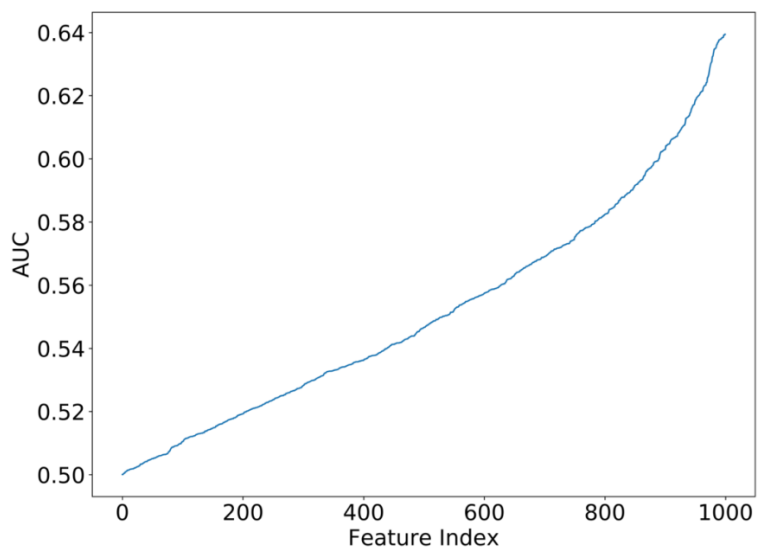

Figure 4. AUCs of each individual feature of the final fully connected layer's feature map using GoogleNet.

## Discussion

We studied the molecular subtype classification of breast cancer using deep learning applied to dynamic contrast-enhanced magnetic resonance imaging. Three different deep learning approaches were investigated, and among them off-the-shelf deep features approach performed best. The highest AUC obtained was 0.65 using pre-trained GoogleNet as feature extractor and a polynomial kernel SVM trained on the extracted features. The highest AUC obtained for training from scratch was 0.58, which is only slightly higher than random guess. The performance of transfer learning is worse than off-the-shelf deep features but better than training from scratch. This implies that when the number of training samples is limited, overfitting is the main problem.

There are two potential ways to solve the overfitting problem. Instead of using pre-training networks on a natural image dataset, using pre-training networks on large medical image dataset seems more promising, as medical images share much more in common. Another way to deal with data insufficiency is to train a model that can do multiple tasks[28]. Tasks that have insufficient training data will benefit from tasks that have sufficient training data. We consider these potential solutions as future work.

The overall prognostic power of our results was low to intermediate with the highest AUC of 0.65. While direct comparison to results of other studies that used hand-crafted features is difficult since the specific goals of different studies, the datasets, and the evaluation metrics vary widely, the obtained level of performance is in the expected range. A direct comparison of deep learning with hand crafted features will be a part of future work.

Our study had some limitations. The most significant one, discussed throughout the paper, is the relatively low number of cases in the context of deep learning. Our study explores this issue and suggests the deep learning solutions that are most likely to succeed in this imperfect situation. It is encouraging to see that even with such a small number of cases we were able to develop a deep learning model that showed some predictive value of molecular subtypes. Another limitation was considering only the distinction between the Luminal A subtype versus all other subtypes, as opposed to considering each molecular subtypes individually. This was due to the small number of non-luminal A cases.

In conclusion, in this study, we were able to demonstrate that deep learning can aid the investigation of the relationship between cancer imaging and tumor imaging in breast cancer. Future studies will include repeating this investigation with a larger dataset and comparison of deep learning to traditional machine learning models based on hand-crafted features.